\documentclass[conference]{IEEEtran}
\IEEEoverridecommandlockouts

\usepackage{cite}
\usepackage{amsmath,amssymb,amsfonts}
\usepackage{algorithmic}
\usepackage{graphicx}
\usepackage{float}  
\usepackage{array}  
\usepackage{textcomp}
\usepackage{xcolor}
\def\BibTeX{{\rm B\kern-.05em{\sc i\kern-.025em b}\kern-.08em
    T\kern-.1667em\lower.7ex\hbox{E}\kern-.125emX}}
\begin{document}

\title{Machine Learning-Based Classification of Oils Using Dielectric Properties and Microwave Resonant Sensing\\

}
\author{
\IEEEauthorblockN{Amit Baran Dey}
\IEEEauthorblockA{\textit{Dept. of EEE, IIT Guwahati,} \\
Assam, India \\
deyamit@iitg.ac.in}
\and
\IEEEauthorblockN{Wasim Arif}
\IEEEauthorblockA{\textit{Dept. of ECE, NIT Silchar,} \\
Assam, India \\
arif@ece.nits.ac.in}
\and
\IEEEauthorblockN{Rakhesh Singh Kshetrimayum}
\IEEEauthorblockA{\textit{Dept. of EEE, IIT Guwahati,} \\
Assam, India \\
krs@iitg.ac.in}
}

\maketitle

\begin{abstract}
This paper proposes a machine learning-based methodology for the classification of various oil samples based on their dielectric properties, utilizing a microwave resonant sensor. The dielectric behavior of oils, governed by their molecular composition, induces distinct shifts in the sensor’s resonant frequency and amplitude response. These variations are systematically captured and processed to extract salient features, which serve as inputs for multiple machine learning classifiers. The microwave resonant sensor operates in a non-destructive, low-power manner, making it particularly well-suited for real-time industrial applications. A comprehensive dataset is developed by varying the permittivity of oil samples and acquiring the corresponding sensor responses. Several classifiers are trained and evaluated using the extracted resonant features to assess their capability in distinguishing between oil types. Experimental results demonstrate that the proposed approach achieves a high classification accuracy of 99.41\% with the random forest classifier, highlighting its strong potential for automated oil identification. The system’s compact form factor, efficiency, and high performance underscore its viability for fast and reliable oil characterization in industrial environments.
\end{abstract}

\begin{IEEEkeywords}
Machine learning, oil classification, dielectric properties, microwave resonant sensor, material characterization.
\end{IEEEkeywords}

\section{Introduction}
In recent years, the study of dielectric permittivity of various materials at microwave frequencies has garnered significant attention within the RF and microwave engineering community. Leveraging the interaction between electromagnetic (EM) energy and material media, a wide array of electronic sensor designs has been developed. Among these, microwave-based techniques have emerged as particularly attractive for material characterization, owing to their ability to provide rapid, non-destructive evaluations with minimal power consumption and compact circuit designs [1]–[5].

Resonant methods, where a sample under test (SUT) interacts with a resonant structure, have proven especially effective. These methods are preferred for their high precision and the straightforward extraction of dielectric properties directly from shifts in the resonant characteristics of the system [2]–[5].

Recent research efforts have begun exploring the use of microwave sensors for testing biofuel mixtures [6]. In this context, machine learning techniques, such as the multilayer perceptron (MLP) algorithm, have been introduced as powerful tools for identifying oil mixtures, addressing uncertainties arising from material structure [6]. This has given rise to a new research direction combining microwave measurement techniques with machine learning models. Furthermore, various microwave-based methods have integrated machine learning to enhance measurement accuracy and reduce computational modeling demands [7]–[9]. Despite these advancements, minimizing measurement uncertainties remains a critical challenge. The use of multiple classification algorithms has shown promise in improving the accuracy of material identification while maintaining simplicity, non-destructiveness, speed, and energy efficiency.

Over the past decade, microwave split-ring resonators (SRRs) have attracted considerable interest for sensing applications, driven by their low fabrication costs, moderate to high sensitivity, high quality factors, compatibility with CMOS technologies, and non-invasive, non-contact operation. Owing to these attributes, SRRs have been successfully employed across diverse industries, including biomedical sensing and the energy sector [10]–[17]. SRRs have demonstrated capabilities such as non-invasive detection of extremely low concentrations of biomarkers like glucose and lactate, and robust analysis of bitumen samples under extreme conditions [18]–[22]. Additionally, they have found applications in liquid sensing, binary detection, and thickness measurements [23]–[31].

Typically, a microwave resonator-based sensor consists of a microstrip ring with a single split, creating a localized region of high electric field intensity, ideal for sensitive material detection. The fundamental sensing mechanism relies on shifts in the resonant frequency caused by variations in the relative permittivity of the material under test. The machine learning training process from the microwave dataset is illustrated in Fig. 1.
\begin{figure}[h]
\centerline{\includegraphics[width = 1\linewidth, scale=1]{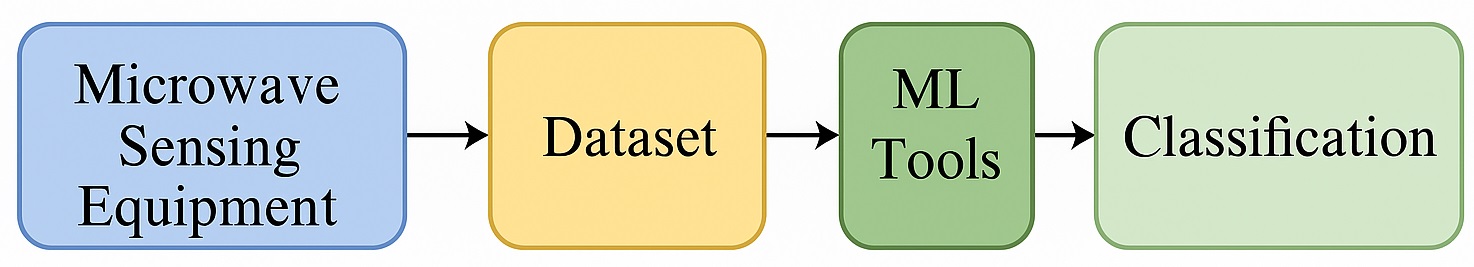}}
\caption{Process for Designing and Preparing a Microwave Dataset for Machine Learning (ML) Training in Sensing-Based Classification.}
\label{fig}
\end{figure}

In this study, we focus on the classification of different oil samples based on their dielectric properties. Traditional single-variable microwave sensors are often insufficient for accurately distinguishing between closely related oil types. To overcome this limitation, we propose a significant enhancement: improving the selectivity of microwave resonator sensors by extracting additional independent features from higher-order harmonics of the resonator’s response. This approach exploits the intrinsic nonlinearity of the permittivity spectrum of oils, which arises from their distinct molecular relaxation behaviors under identical environmental conditions. By analyzing frequency shifts and variations in resonance amplitude across the first two harmonic modes, a comprehensive feature set is developed for accurate oil classification.

\section{Design and Simulation of the Microwave-Based Sensor}

This study utilizes a microstrip-based split-ring resonator (SRR) designed to operate at its fundamental resonant frequency. The resonator structure is based on a planar feed with two excitations placed on opposite sides of the ports. It is assumed to be fabricated on an FR4 substrate, with a thickness of 1.6~mm and a copper thickness of 0.035~mm. FR4 is selected due to its low cost and wide availability. The resonator design is depicted in Fig.~2, with its dimensional parameters (in mm) listed in Table~I.

\begin{figure}[h]
\centerline{\includegraphics[width = 1\linewidth, scale=1]{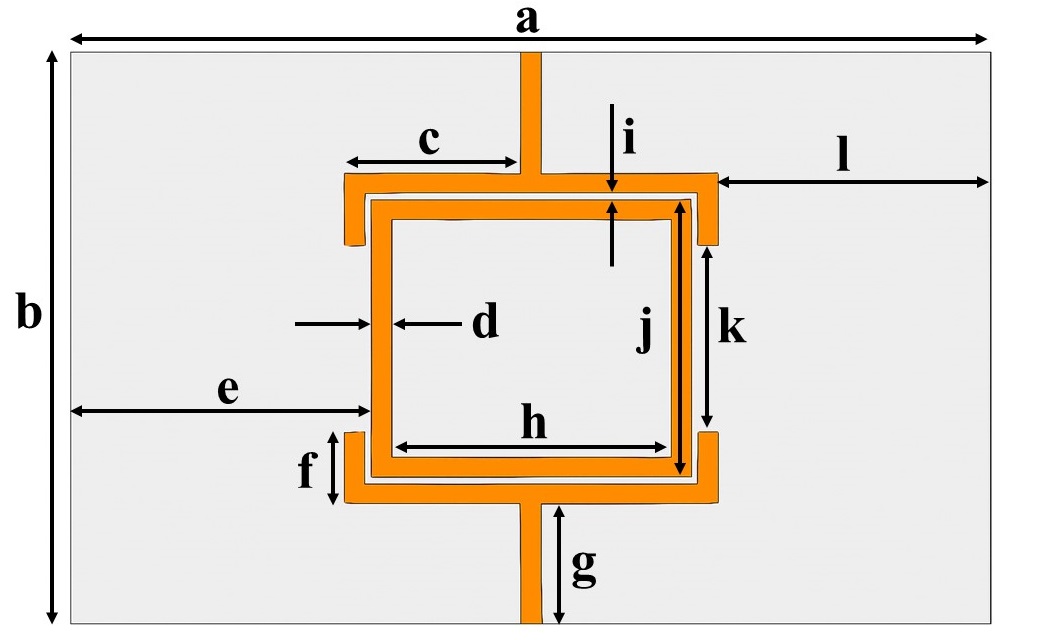}}
\caption{Schematic of the Proposed Microwave Resonant Sensor.}
\label{fig1}
\end{figure}

The simulated transmission coefficient ($S_{21}$) response of the proposed microwave resonant sensor, shown in Fig.~3, is analyzed across a frequency range of 1--4~GHz. Two distinct resonant peaks are observed, corresponding to the first and second harmonics of the sensor. The first resonance occurs around 1.4~GHz, and the second near 2.8~GHz. Both resonances exhibit clear dips in the $S_{21}$ response, indicating strong coupling and effective energy transmission at the designed frequencies.
\begin{table}[h!]
\centering
\caption{Dimensional Parameters of the Proposed Microwave Resonant Sensor}
\label{tab:bare_resonator}
\begin{tabular}{|l|l|l|l|}
\hline
\textbf{Parameter} & \textbf{Values} & \textbf{Parameter} & \textbf{Values}\\
\hline
\textbf{a} & \text{110}
 & g  & 15.7 \\
 \hline
\hline\textbf{b} & \text{70}
 & h  & 30 \\
 \hline
\hline\textbf{c} & \text{18.35}
 & i  & 0.5 \\
 \hline
 \hline\textbf{d} & \text{1.9}
 & j  & 33.8 \\
 \hline\hline\textbf{e} & \text{38.1}
 & k  & 23 \\
 \hline
\hline\textbf{f} & \text{7.8}
 & l  & 35.7 \\
 \hline
\end{tabular}
\end{table}

\begin{figure}[h]
\centerline{\includegraphics[width = 1\linewidth, scale=1]{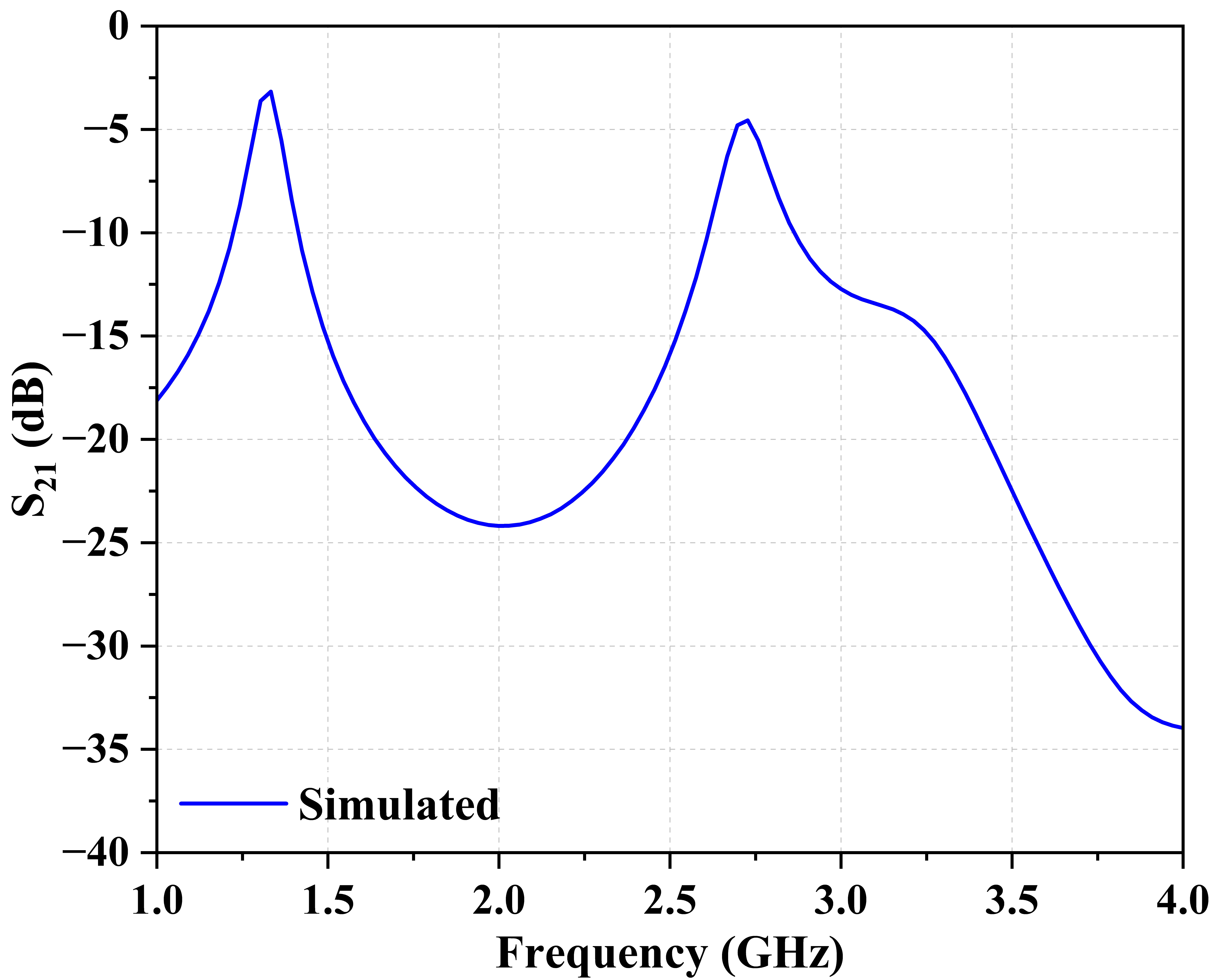}}
\caption{Transmission Characteristics of the First Two Harmonics of the Proposed Microwave Resonant Sensor.}
\label{fig3}
\end{figure}

The sharpness of these resonances underscores the sensor’s high-quality factor, making it well-suited for dielectric characterization applications. The low insertion loss in the passbands further demonstrates the efficiency of the resonator structure. At these frequencies, the dielectric properties of the oil samples under investigation show significant variations, especially at the higher-order harmonics. The core objective of this study is to identify frequency shifts across the first two harmonics, alongside corresponding changes in transmission amplitude and quality factors, which serve as distinctive features for classifying different oil types based on their molecular composition.

\section{Simulation Setup}

Fig. 4 illustrates the simulation setup of the proposed microwave resonant sensor used for oil classification. The resonator is positioned at the bottom of the oil sample, with the distance between the resonator and the oil interface varying along the z-axis from 1 mm to 50 mm. For machine learning model training, finer variations along the z-direction, starting from Z = 0.001 mm, are considered to generate a large dataset. The oils analyzed in this study include soybean oil, olive oil, coconut oil, and peanut oil, with their dielectric properties assigned based on standard values reported in the literature [34--37].

As shown in Fig.~5 -- Fig.~8, the loading of the resonator with olive, peanut, soybean, and coconut oil demonstrates that each material exhibits unique nonlinear behaviors in both permittivity and conductivity at the resonator’s harmonic frequencies. These frequency-dependent variations arise from differences in relaxation times across the materials [32--33], introducing novel features that enhance the sensor’s capability for multi-oil classification.

\begin{figure}[h]
\centerline{\includegraphics[width = 1\linewidth, scale=1]{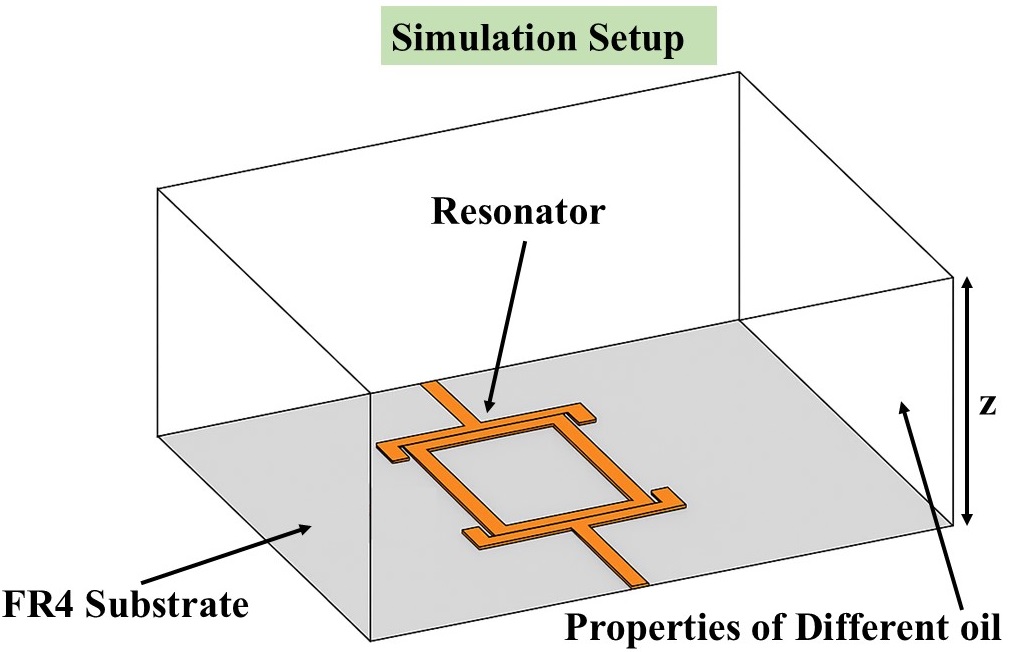}}
\caption{Simulation Setup of the Proposed Microwave Resonant Sensor for Oil Classification.}
\label{fig:setup}
\end{figure}

\begin{figure}[h]
\centerline{\includegraphics[width = 1\linewidth, scale=1]{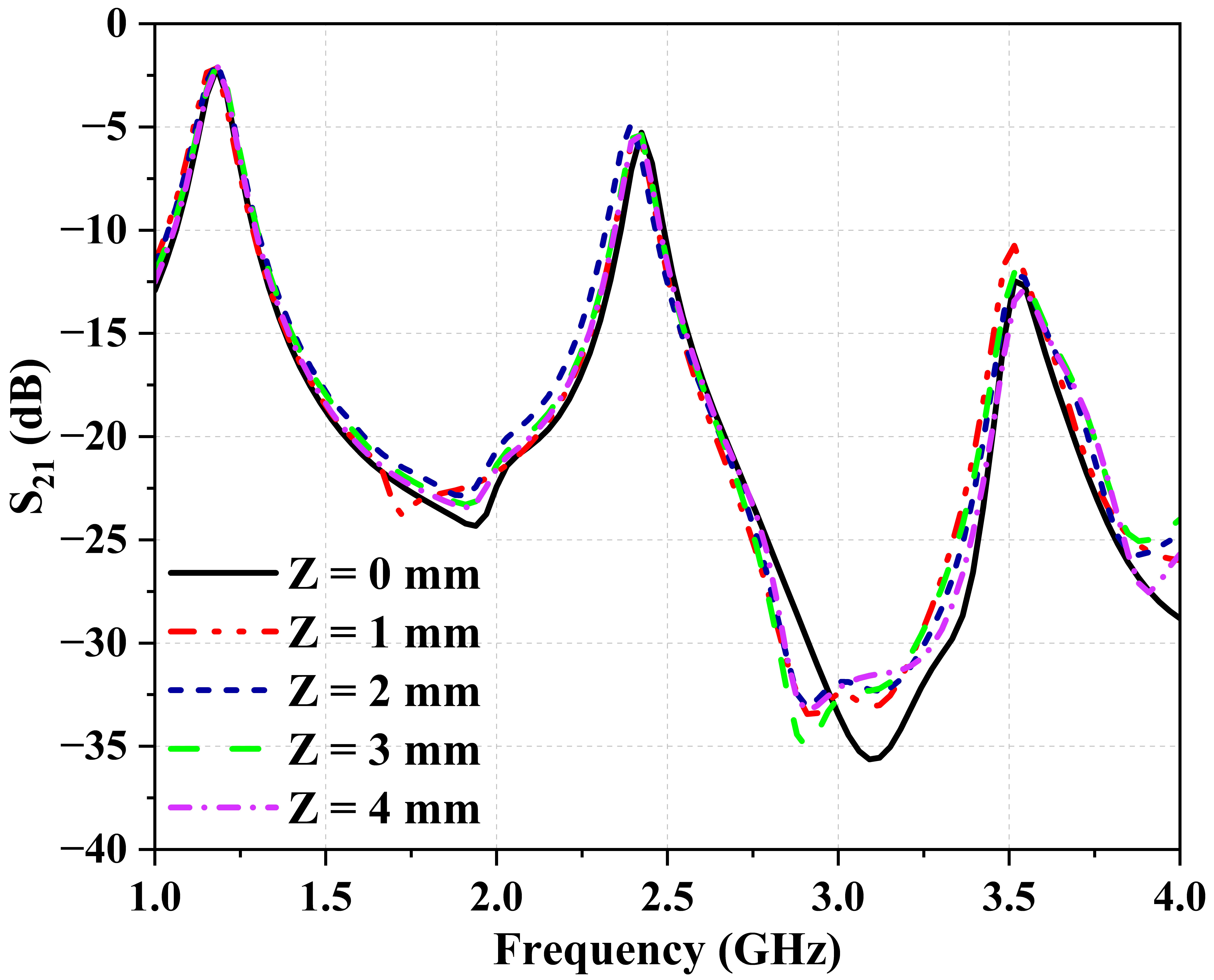}}
\caption{Transmission Characteristics of the First Two Harmonics of the Proposed Microwave Resonant Sensor with Olive Oil Loading.}
\label{fig:transmission}
\end{figure}

Some harmonics, where permittivity remains constant across materials, do not contribute independent features and are therefore considered redundant. In contrast, harmonics that yield distinct permittivity values increase the sensor’s discriminatory power, enabling more accurate multi-oil classification and material characterization using microwave SRR-based platforms. To enrich the feature space and improve training accuracy for the artificial neural network (ANN), the first two resonance modes were selected for feature extraction, where shifts in resonance frequencies under different oil loadings provide key discriminatory information. These extracted features are discussed further in the data analysis section.

In the context of oil analysis, Fig.~5 presents the transmission spectra of the resonator under olive oil loading condition. A noticeable shift in resonance frequency is observed as the oil loading increases, highlighting the resonator's sensitivity to material properties. The operating range from 1~GHz to 4~GHz encompasses multiple resonances, with significant permittivity variations among the oils, making this range ideal for characterization. The resonator was initially optimized to resonate at approximately 1.45~GHz in its fundamental mode under unloaded conditions ($Z=0$~mm).

\label{fig5}
\begin{figure}[h]
\centerline{\includegraphics[width = 1\linewidth, scale=1]{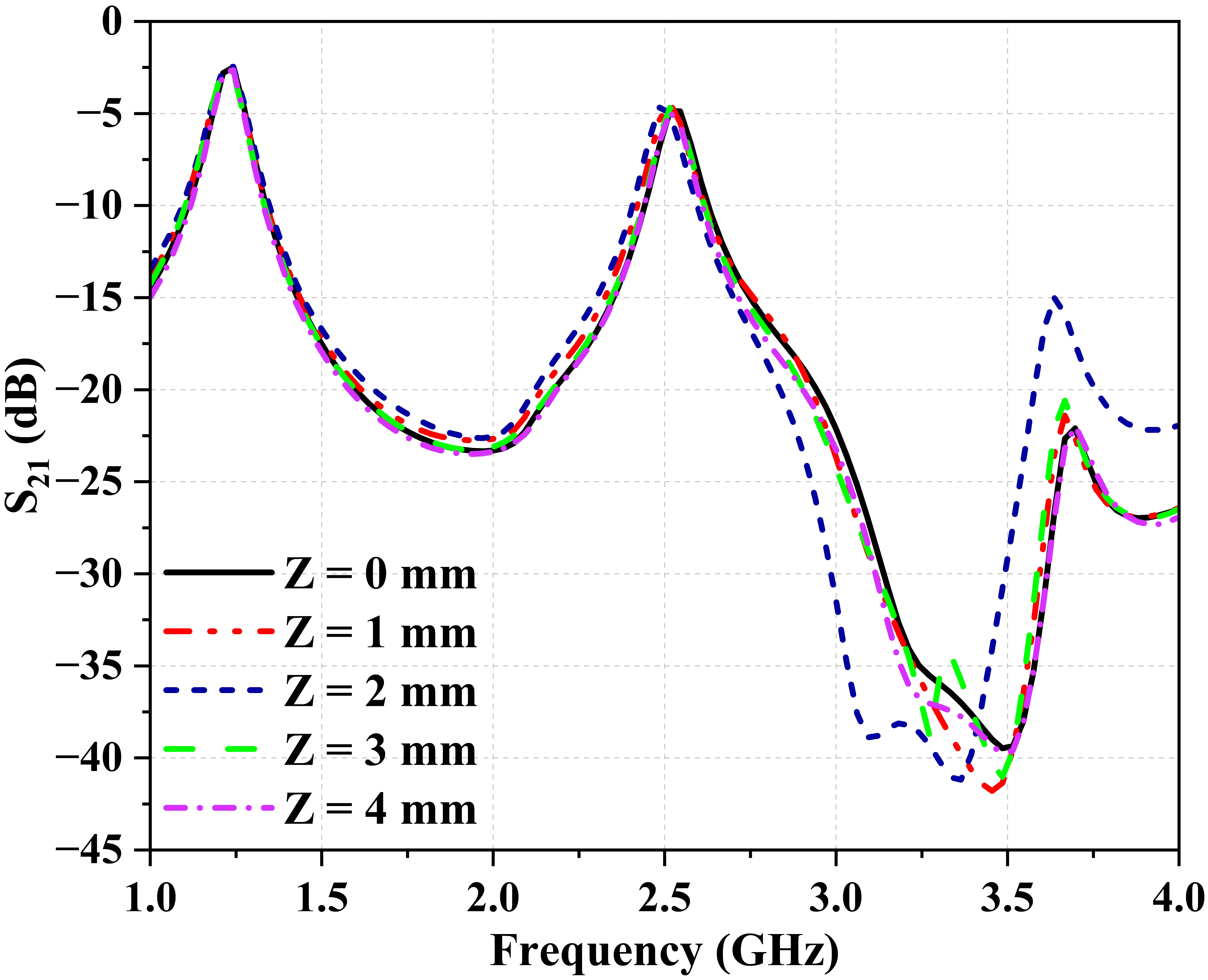}}
\caption{Transmission Characteristics of the First Two Harmonics of the Proposed Microwave Resonant Sensor with Peanut Oil Loading.}
\end{figure}

\label{fig6}
\begin{figure}[h]
\centerline{\includegraphics[width = 1\linewidth, scale=1]{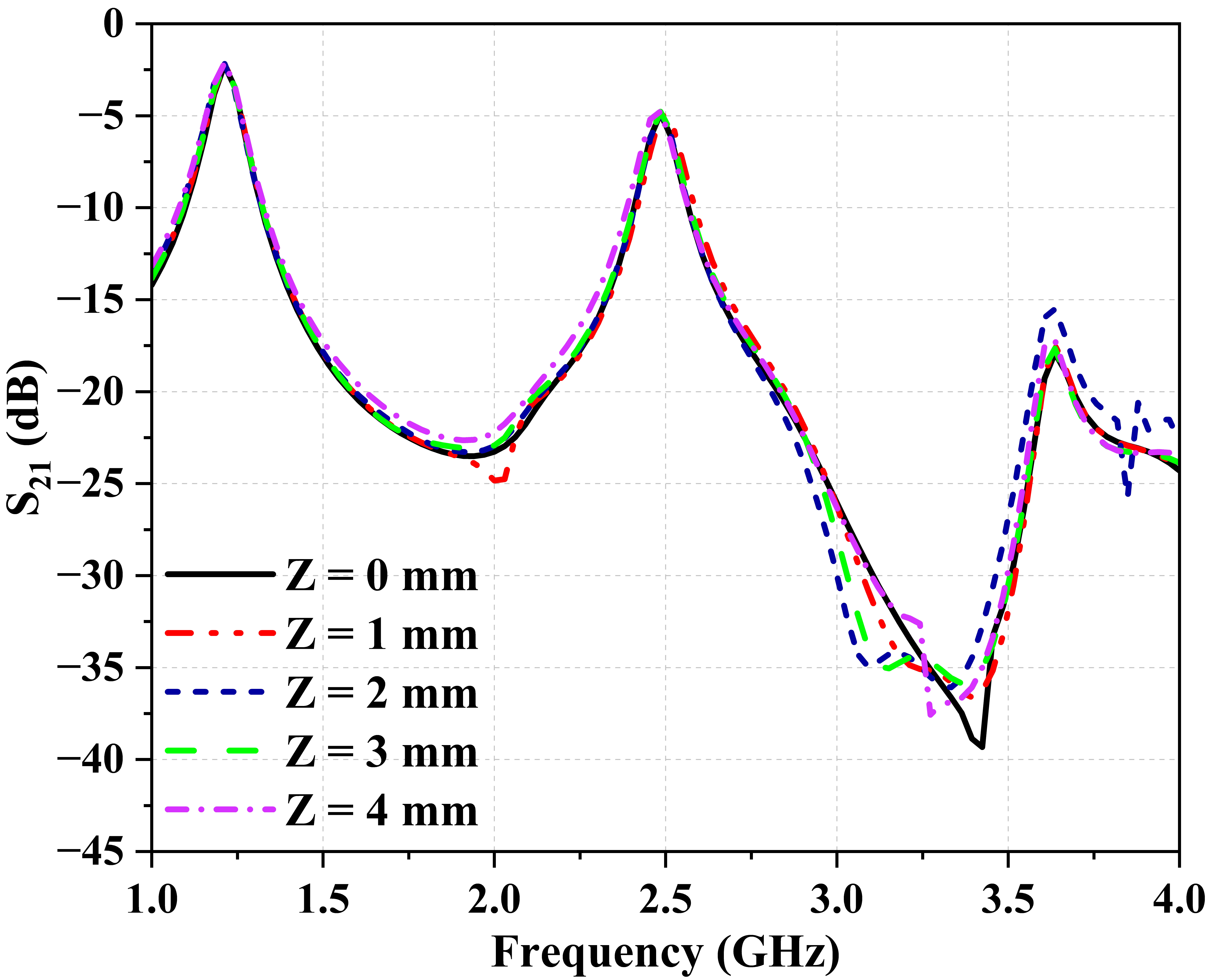}}
\caption{Transmission Characteristics of the First Two Harmonics of the Proposed Microwave Resonant Sensor with Soybean Oil Loading.}
\label{fig6}
\end{figure}

Fig.~6 displays the schematic of the designed resonator and its simulated transmission spectrum, focusing on the first two harmonics with peanut oil loading. The field distribution patterns across the harmonics remain nearly uniform, with shifts due to the use of air as the ambient medium during simulations. To assess feature independence, normalized frequency shifts at different harmonics are analyzed relative to the unloaded base resonance frequency.

\begin{figure}[h]
\centerline{\includegraphics[width = 1\linewidth, scale=1]{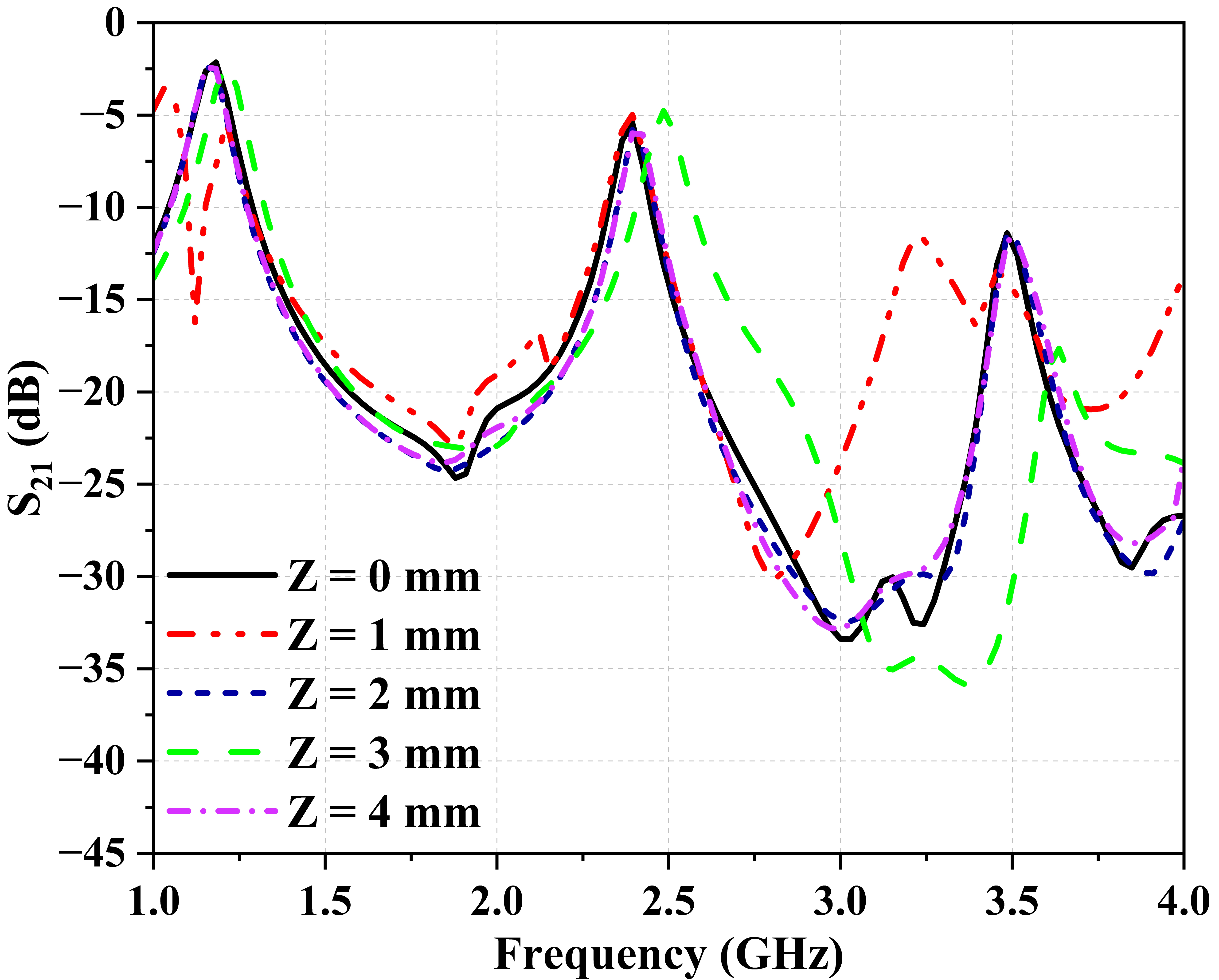}}
\caption{Transmission Characteristics of the First Two Harmonics of the Proposed Microwave Resonant Sensor with Coconut Oil Loading.}
\label{fig7}
\end{figure}

Similarly, the transmission coefficient $S_{21}$ was measured for different distances ($Z$) between the sensor and the reference medium, using soybean oil and  coconut oil as shown in Fig.~7 and Fig.~8 respectively. As observed, the resonance frequencies remain relatively stable, with slight variations in magnitude as $Z$ increases from 0~mm to 4~mm, indicating the sensor’s robustness without significant frequency shifts.

\section{Machine Learning-Assisted Oil Analysis Using a Microwave Resonant Sensor}

Machine learning algorithms were employed to enhance the analysis of transmission response data from the microwave resonant sensor. Several classifiers, including Logistic Regression, $k$-Nearest Neighbors (KNN), Random Forest, and Support Vector Machine (SVM), were evaluated based on their performance metrics: accuracy, precision, recall, and F1 score.

The dataset comprises measured scattering parameters (S-parameters), specifically focusing on the \( S_{21} \) transmission coefficient, across varying frequencies and heights for four different oil types: coconut Oil, olive Oil, peanut Oil, and soyabean Oil. The \( S_{21} \) parameters represent the amplitude variation (in dB) as a function of frequency at different sample heights, providing distinctive dielectric signatures for each oil.

In the implemented code:

\begin{itemize}
    \item Data preprocessing involved standardizing column names (height, frequency, \( S_{21} \)), assigning numerical labels to oil types, and removing missing or duplicate entries.
    \item The feature set includes height, frequency, and \( S_{21} \), while the target variable is the oil class label.
    \item Features are normalized using standardization (zero mean, unit variance).
    \item The dataset is split into 80\% training and 20\% testing subsets.
\end{itemize}

Four machine learning models were developed and compared for classification:
\begin{enumerate}
    \item Random Forest Classifier
    \item Support Vector Machine (SVM)
    \item Logistic Regression
    \item k-Nearest Neighbors (k-NN)
\end{enumerate}

Model performances were evaluated based on accuracy, confusion matrix, and classification reports. This approach leverages the variation of \( S_{21} \) amplitude with frequency to enable accurate classification of different oils.

As shown in Fig.~9, KNN, Random Forest, and SVM demonstrated high and comparable performance across all metrics, while Logistic Regression showed significantly lower accuracy and recall. These findings emphasize the effectiveness of advanced machine learning models in ensuring accurate classification in oil analysis applications.

Fig.~10 presents a comparative analysis of the classification performance of the four machine learning models---Logistic Regression, $k$-Nearest Neighbors (KNN), Random Forest, and Support Vector Machine (SVM) based on key evaluation metrics: accuracy, precision, recall, and F1 score. Logistic Regression exhibits relatively poor performance, achieving approximately 53.52\% accuracy. Its precision and recall values are also low, indicating difficulty in correctly identifying and distinguishing between oil classes. While the F1 score for Logistic Regression (0.62) is slightly higher than its individual precision and recall, it remains significantly lower than those of the other classifiers. This suggests that Logistic Regression may be unsuitable for handling complex, non-linear distributions commonly found in oil sample classification tasks.

\begin{figure}[h]
\centerline{\includegraphics[width = 0.94\linewidth, scale=1]{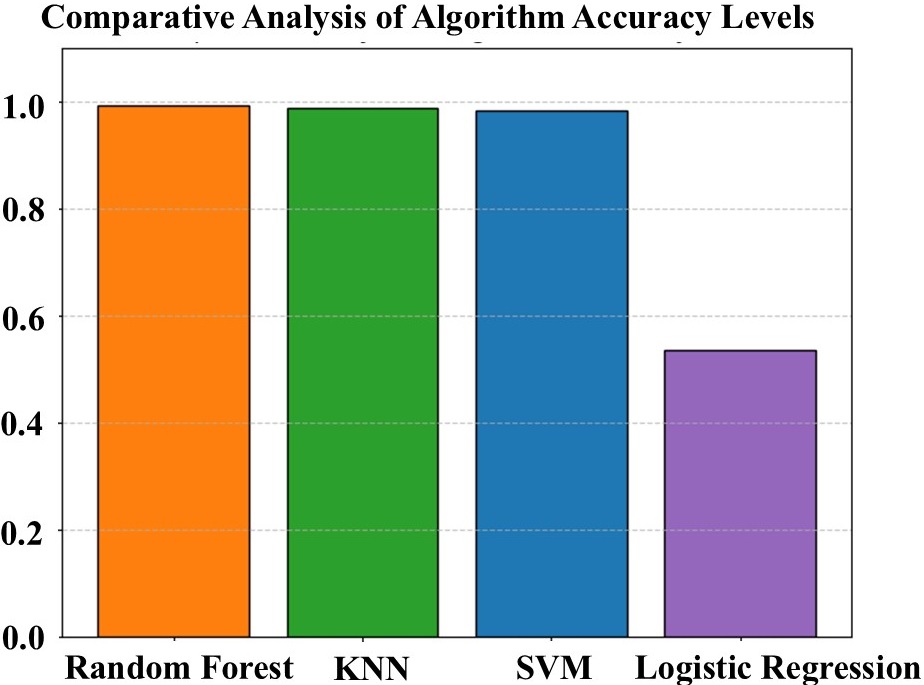}}
\caption{Evaluation of Algorithm Accuracy Levels.}
\label{fig8}
\end{figure}

In contrast, the KNN, Random Forest, and SVM classifiers all achieve accuracies exceeding 98\%, with precision, recall, and F1 scores consistently approaching 1.0. Among these, Random Forest marginally outperforms the others in terms of recall, demonstrating its enhanced ability to correctly identify various oil classes. This superior performance can be attributed to the ensemble nature of Random Forest, which helps mitigate overfitting and manage the high-dimensional feature spaces often associated with oil characterization. 

\begin{figure}[h]
\centerline{\includegraphics[width = 1\linewidth, scale=1]{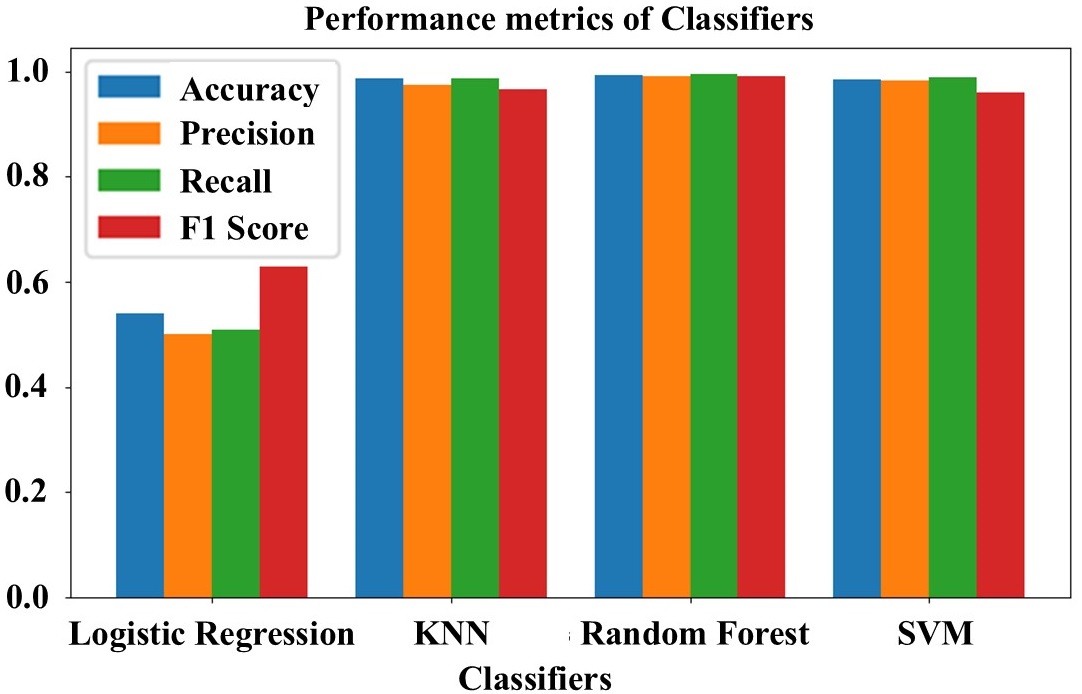}}
\caption{Evaluation of Performance Metrics for All Classifiers.}
\label{fig9}
\end{figure}

Similarly, KNN maintains high metric scores across all evaluation criteria, benefiting from its instance-based learning approach, which is particularly effective when the data distribution is well-separated. The SVM model also shows strong classification performance, although a slight decrease in the F1 score compared to Random Forest and KNN suggests a minor imbalance between precision and recall. Overall, these results indicate that Random Forest is the most effective model for oil classification, achieving near-perfect classification accuracy (99.41\%). In contrast, Logistic Regression is less suitable for this task, likely due to its linear decision boundary limitations, which are not well-suited for the complex and non-linear feature space characteristic of oil sample datasets.

\begin{figure}[h]
\centerline{\includegraphics[width = 0.85\linewidth, scale=1]{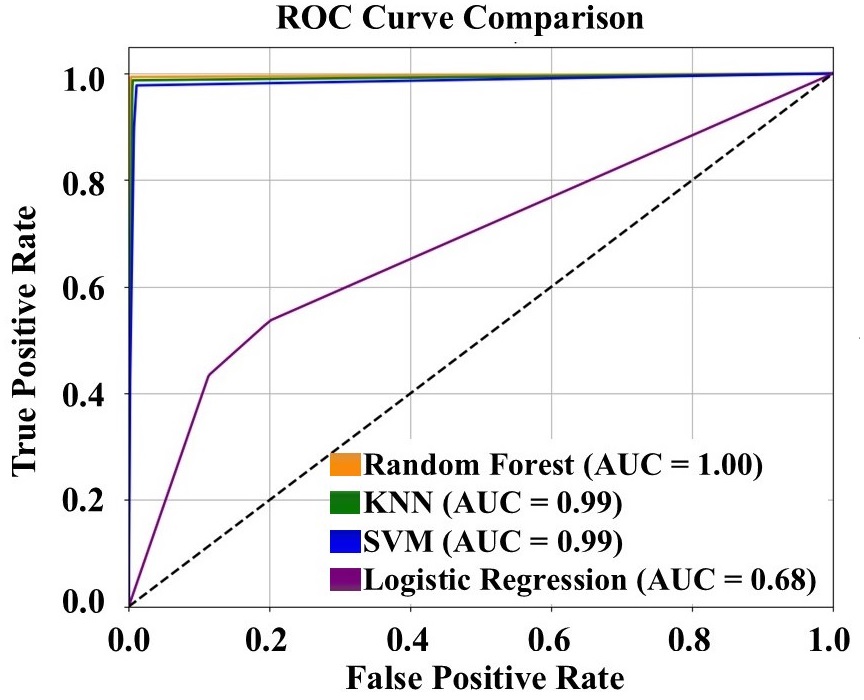}}
\caption{Comparison of ROC Curves for All Considered Models.}
\label{fig10}
\end{figure}

Fig.~11 presents the Receiver Operating Characteristic (ROC) curves for the evaluated classifiers---Random Forest, $k$-Nearest Neighbors (KNN), Support Vector Machine (SVM), and Logistic Regression used in the oil classification task. The Random Forest classifier achieves an AUC (Area Under the Curve) of 1.00, indicating perfect classification performance with no false positives or false negatives. KNN and SVM follow closely with AUC values of 0.99, reflecting their excellent discriminatory capabilities. In contrast, Logistic Regression demonstrates a substantially lower performance, with an AUC of 0.68, suggesting its limited effectiveness in handling the complexity of the oil dataset. The ROC curves clearly show that Random Forest, KNN, and SVM are highly reliable for accurate oil classification, while Logistic Regression is less suitable due to its poor ability to separate classes.

Considering the dielectric characterization of oil samples using a microwave resonant sensor and the subsequent machine learning analysis, it is clear that integrating resonator-based sensing with advanced classification algorithms significantly improves oil identification accuracy. The study demonstrates that resonant frequency shifts reliably capture the material properties, and machine learning classifiers particularly Random Forest, KNN, and SVM successfully interpret these features, achieving accuracies above 98\% and AUC values approaching 1.00. In contrast, Logistic Regression, limited by its linear nature, underperforms, reinforcing the need for more sophisticated models in complex material classification tasks. Overall, the combination of microwave sensing and machine learning provides a fast, non-destructive, and highly accurate methodology for oil classification, with Random Forest emerging as the most effective model.

\section{Conclusion}
This work presents a highly accurate machine learning-based approach for classifying oil samples based on their dielectric properties using a microwave resonant sensor. By analyzing variations in resonant frequency and amplitude response, the system achieves a classification accuracy of 99.41\% with the random forest classifier, outperforming the other evaluated models. The sensor’s non-destructive and low-power operation makes it particularly suitable for real-time industrial applications. The results underscore the promising potential of integrating microwave sensing with machine learning techniques for fast, efficient, and reliable oil characterization, contributing to the advancement of intelligent sensing systems for material analysis.

\end{document}